\documentclass[preprint,12pt]{elsarticle}

\usepackage{amssymb}
\usepackage{amsmath,amssymb,amsfonts}
\usepackage{algorithmic}
\usepackage{algorithm}
\usepackage{graphicx}
\usepackage{textcomp}
\usepackage{verbatim}
\usepackage{multirow}
\usepackage{tabu}
\usepackage{makecell}
\usepackage{color}

\usepackage{wrapfig}
\usepackage{url}

\journal{arXiv}

\begin{document}

\begin{frontmatter}



\title{An Automated Data Engineering Pipeline for Anomaly Detection of IoT Sensor Data}


\author{Xinze Li and Baixi Zou}
\ead{\{xli2966, bzou23\}@uwo.ca}

\address{Department of Electrical and Computer Engineering, University of Western Ontario, 1151 Richmond St, London, Ontario, Canada N6A 3K7}

\begin{abstract}
The rapid development in the field of System of Chip (SoC) technology, Internet of Things (IoT), cloud computing, and artificial intelligence has brought more possibilities of improving and solving the current problems. With data analytics and the use of machine learning/deep learning, it is made possible to learn the underlying patterns and make decisions based on what was learned from massive data generated from IoT sensors. When combined with cloud computing, the whole pipeline can be automated, and free of manual controls and operations. In this paper, an implementation of an automated data engineering pipeline for anomaly detection of IoT sensor data is studied and proposed. The process involves the use of IoT sensors, Raspberry Pis, Amazon Web Services (AWS) and multiple machine learning techniques with the intent to identify anomalous cases for the smart home security system. 
\end{abstract}


\end{frontmatter}


\section{Introduction}
With the development of System of Chip (SoC) technologies, embedded computing chips are becoming increasingly computationally efficient and smaller in size. Engineers can now design and connect devices to the Internet of Things (IoT) \cite{r1} \cite{r2}. At the same time, cloud computing, especially serverless deployment, is becoming the first choice of many companies as it reduces the need to manage hardware and clusters \cite{r3}. Recent progress of machine learning benefits from hardware development and is able to provide better performance with faster speed \cite{r4} \cite{r5} \cite{r6_}. It is also combined with cloud computing technologies leading to results like online data pipeline management platforms. Data generated from IoT-connected sensors can be massive in size. It would require further analysis to transform it into meaningful information and extract insights from the massive dataset \cite{r6}. With data analytics and the use of machine learning/deep learning, it has the ability to learn the underlying patterns and make decisions based on what was learned. Furthermore, with data engineering, the automated data pipeline can free the time spent on triggering the pipeline manually and therefore, ease the maintenance and the whole process \cite{r7}. 

The objective of this project is to design and build an end-to-end data engineering pipeline that can automate ML model training and deployment. Our design considerations include the type of database to use at the edge and at the cloud, data analytics tools used to gain insight into raw data, and how to deploy trained models back to the edge machine. This data engineering pipeline is a generic pipeline that can be used for various IoT data analytics problems. As a typical use case, the pipeline is evaluated on a typical use case: automated anomaly detection on real-world sensor data \cite{r8} \cite{r9}. In the experiments, there are two Raspberry Pi machines located in two different environments in a house. One is placed on the door of the garage and the other one is placed in the bedroom. The modeling part in the pipeline analyzes and identifies special cases in both bedroom and garage environments. Through the data analytics results from both environments, it can be used as an anti-theft security system for smart homes \cite{r10}. 

The pipeline starts with data generated from IoT sensors which are uploaded to a database with Raspberry Pi and AWS IoT Core \cite{r11}. Then the data will be retrieved from the database and preprocessed for modeling. Machine learning models such as Isolation Forest \cite{r12} and K-means clustering \cite{r13} are applied to identify the anomalous cases wanted. 

The paper covers the following sections: Section II introduces the background of the project including the tools and Amazon Web Services used. Section III introduces the flowchart of the proposed framework as well as the pros and cons of the architecture and its components. Section IV describes data generation (using Raspberry Pi and sensor modules), data transmission (extracting/loading data from database), data preprocessing (removing unnecessary rows, converting feature format, feature encoding), and modeling (Isolation Forest, K-means). Finally, all the results will be displayed and further discussed in Section V. All the scripts for modeling are written in Python 3; and the Python packages, such as pandas, sklearn, and matplotlib, are used for different purposes.

\section{Background}
This section will introduce the tools and services used to build the automated data pipeline and the pros and cons of using them. It involves several Amazon Web Services including AWS IoT core, S3, DynamoDB, PostgreSQL, IoT Analytics, Greengrass, and Local Lambda Functions.

\subsection{Amazon Web Services}

AWS IoT Core \cite{r14} is a service offered by Amazon Cloud that allows connected devices to interact with AWS cloud applications and other devices securely. AWS IoT Core runs on AWS’s serverless architecture and can take the full benefit of the elasticity of the AWS cloud by scaling horizontally to adjust for resource demands. AWS IoT Core supports various protocols, such as HTTPS, LoRaWAN (Long Range Wide Area Network), and MQTT (Message Queuing and Telemetry Transport) protocols. 
AWS S3 (Amazon Simple Storage Service) \cite{r15} is an object storage service offered by Amazon Cloud with high redundancy . It is organized in a folder structure for easy organization. S3 can integrate with other AWS cloud services such as Lambda and SNS for event-driven applications. For long-term cold data storage, developers can implement lifecycle policies in S3 to reduce costs for storing large amounts of data. 

Amazon DynamoDB \cite{r16} is a fully managed NoSQL database service offered by AWS. It can guarantee millisecond latency for applications at any scale. In this project, we use DynamoDB to store and retreat processed sensor data when training anomaly detection models.   

AWS IoT Analytics \cite{r17} is a fully managed service provided by AWS to run data analytics on massive amounts of IoT sensor data. It can automate steps to filter, enrich and transform time-series data. In this project, we pair IoT Analytics with Quicksight to visualize data samples before running the Sagemaker pipeline.  

AWS IoT Greengrass \cite{r18} is managed cloud services provided by AWS to build, deploy and manage embedded device software. In this project, we use AWS Greengrass to manage the software in raspberry pi. 

AWS Lambda \cite{r19} is a serverless computing service provided by AWS. With Greengrass, we can deploy lambda functions directly into the embedded devices, in our case, it would be a raspberry pi. For our use case, we deployed two lambda functions into the raspberry pi. The first one is the software that reads and sends data to the AWS cloud. The second one is the trained anomaly detection model. 

\subsection{Databases (DynamoDB vs. PostgreSQL)}
DynamoDB and PostgreSQL are two different databases provided as Amazon Web Service that are widely used by corporations and individuals \cite{r20}. For this project, PostgreSQL was first used for testing purposes and then converted into DynamoDB as the final choice. Although both databases worked, it is necessary to select the better option based on their differences in features.

DynamoDB is a NoSQL database designed for high performance, scalability, durability, flexibility, and cost-efficiency. PostgreSQL, on the other hand, is a relational database that is simple to set up and use. For this project, all the sensor data will be written into a single table. There are no relations between different tables and the size could be large as new data are constantly being pushed into the database. Additionally, processing large-scale data requires a fast reading speed. According to Amazon DynamoDB’s Developer Guide, DynamoDB has high availability and fast performance as it automatically spreads the data over a sufficient number of servers and is stored on Solid-State Drives. As a result, we decided to use DynamoDB as the data warehouse for this project.
\subsection{Apache Airflow}
Apache Airflow \cite{r21} is an open-source workflow management system and platform that allows users to programmatically author, schedule, and monitor workflows as Directed Acyclic Graphs (DAGs) of tasks \cite{r22}. Airflow Scheduler executes the tasks based on the DAG and schedule defined or external triggers.
\section{System Architecture Design}
\subsection{Architecture for Data Processing v.1 (PostgreSQL, sample data, and Airflow)}
 
 \begin{figure}
     \centering
     \includegraphics[width=8.5cm]{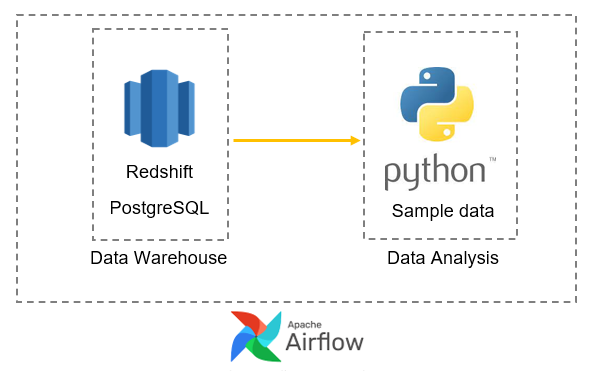}
     \caption{Data Processing Architecture v.1.} \label{iot}
\end{figure}

The initial step of the project aims to build a simple pipeline for data processing. As Figure 1 shows,  the pipeline uses a sample dataset, a PostgreSQL relational database, and Apache Airflow. The sample data was downloaded from DS2OS traffic traces, Kaggle \cite{r23} and was loaded in a PostgreSQL server on AWS Redshift. Through the python script, all the data stored in the database are retrieved, the model is trained and tested, and four evaluation scores are returned as the output. The process is automated by Apache Airflow and is automatically triggered based on the time interval setting.

\subsection{Data Pipeline Architecture v.2}
As the project progressed, we found that our sample data structure had to change to remain aligned with our project requirement. However, because PostgreSQL is a relational database, changing database structure and migrating old data with the new sample data can not be done easily. Hence, the initial data processing architecture is upgraded to v2 as Figure 2 shows.

 \begin{figure}
     \centering
     \includegraphics[width=13cm]{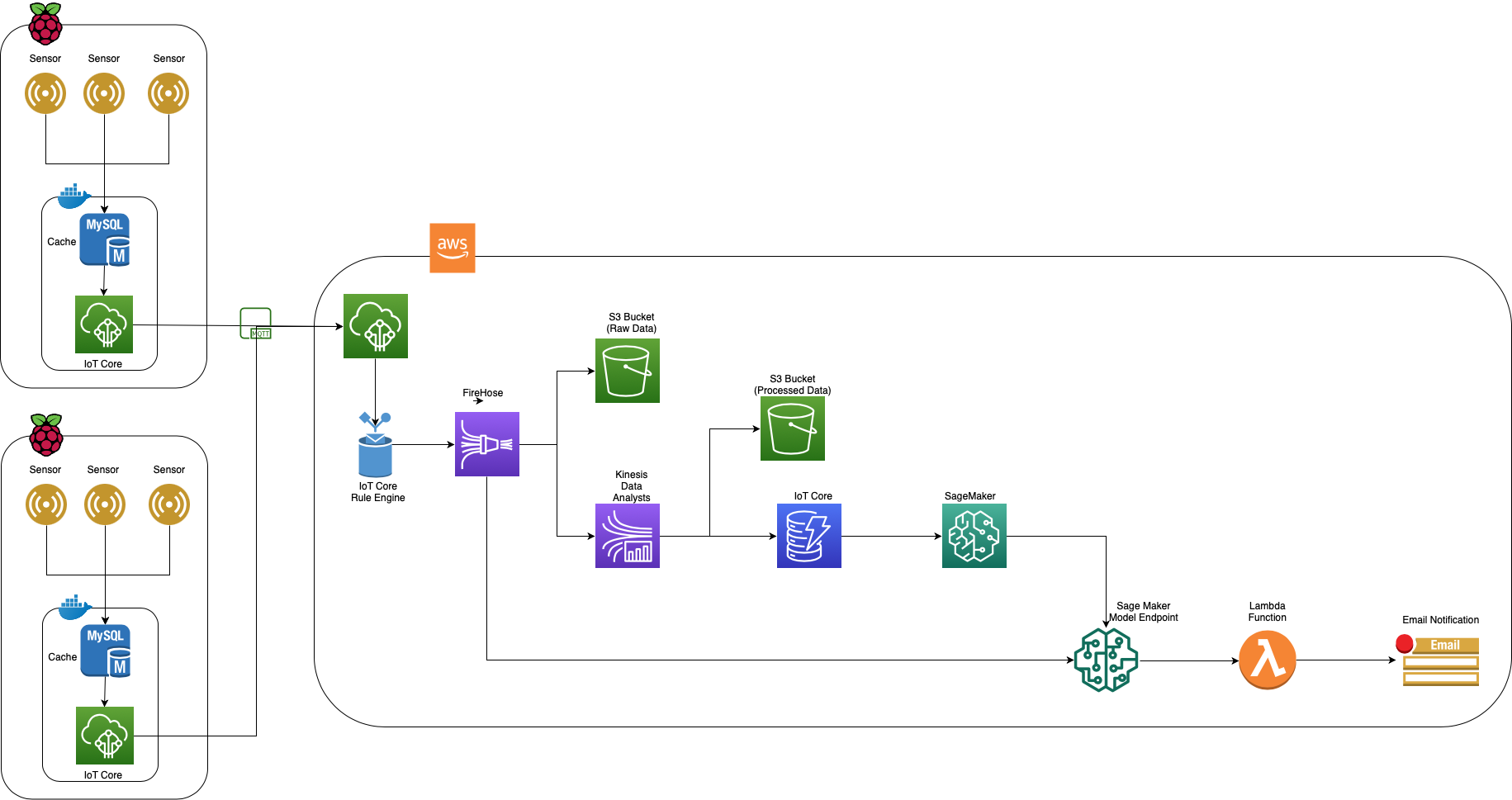}
     \caption{Data Pipeline Architecture v.2.} \label{iot}
\end{figure}

This architecture is inspired by the AWS cloud solution. Data generated in the raspberry pi will first be stored in a local MySQL cache database to avoid sample loss due to unstable internet. Then we use IoT Core to send sample data to the AWS cloud securely. In the AWS cloud, we use Kinesis Firehose to send raw data to a S3 bucket for backup, and to Kinesis Data Analytics for processing. Another S3 bucket is used to back up the processed data before storing the processed data to a DynamoDB table. Every three days, a Sagemaker pipeline will scan the entire DynamoDB table for sample data and run the pipeline to create a model. This model will create a model endpoint that can take sample data from the firehose for real-time anomaly detection. If an anomaly data sample is found, a lambda function will be invoked to send an email notification containing the anomaly data sample. 
\subsection{Data Pipeline Architecture v.3}
 The second architecture design works fine, but we found that it is expensive to run Kinesis Data Analytics. Besides, our data pipeline is not real-time critical, and we can allow for some delay in data processing as the Sagemaker pipeline only runs once every three days. Hence, we changed from using AWS Kinesis to IoT Data Analytics to reduce cost. At the same time, we also integrated Quicksight into the architecture so we can take a quick look at the data sample. Figure 3 displays the third version of the architecture.  
 
  \begin{figure}
     \centering
     \includegraphics[width=13cm]{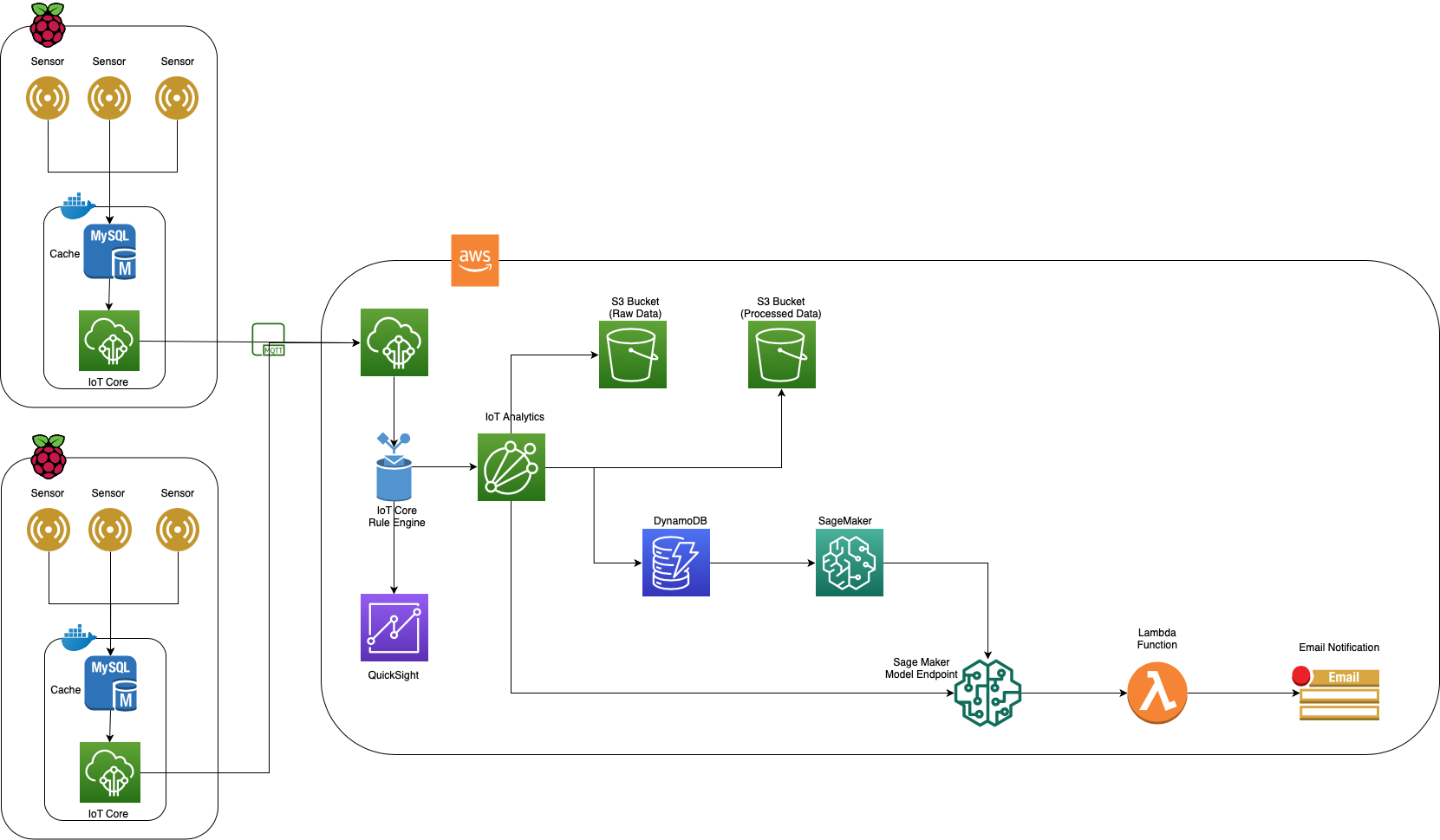}
     \caption{Data Pipeline Architecture v.3.} \label{iot}
\end{figure}

\subsection{Data Pipeline Architecture v.4}
  As our project moves forward, we would like to have real-time anomaly detection at the edge. To satisfy this new requirement, an appropriate way is required to deploy the trained ML model to the raspberry pi. Hence the fourth iteration of the architecture is shown in Figure 4. 
 
  \begin{figure}
     \centering
     \includegraphics[width=13cm]{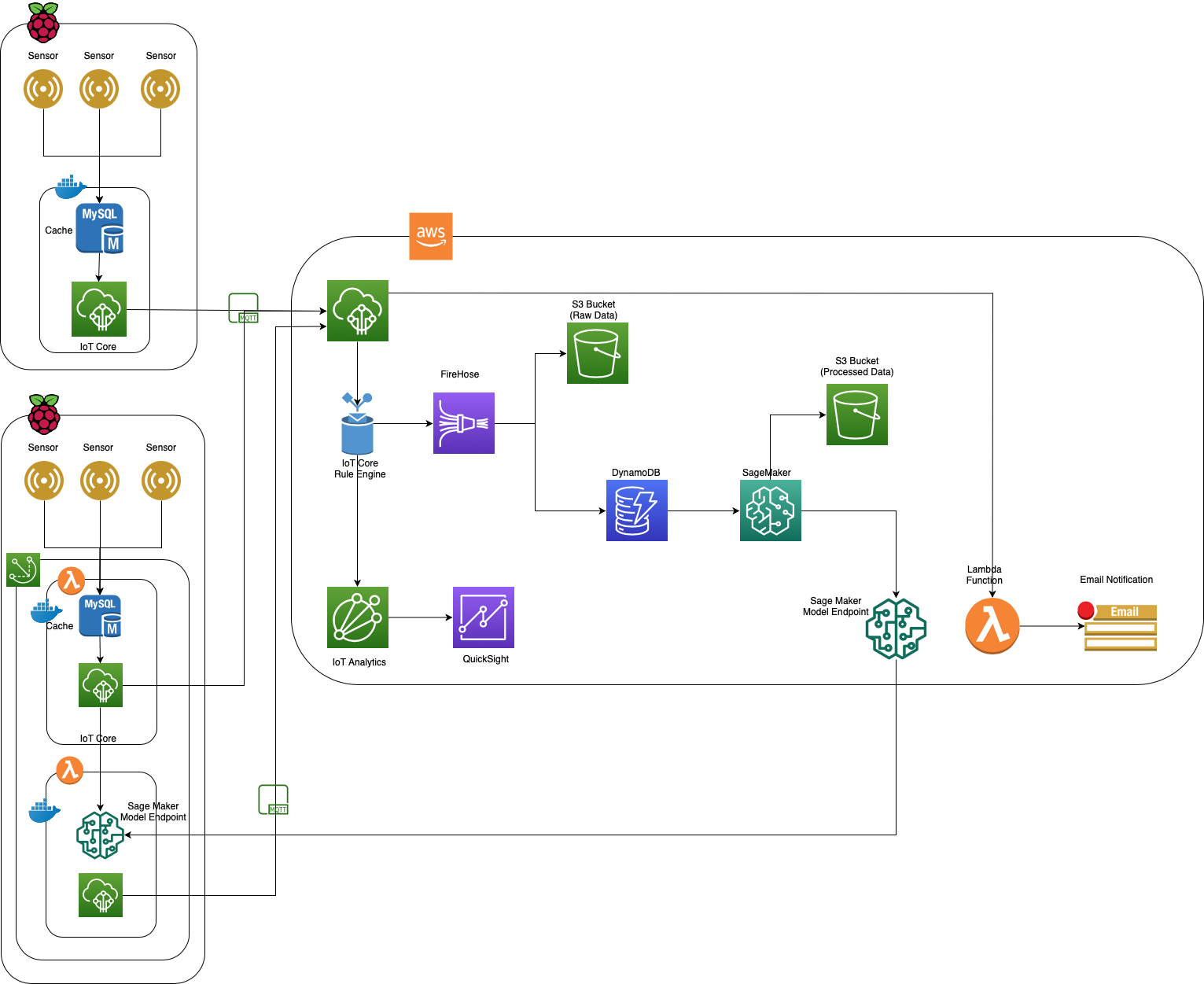}
     \caption{Data Pipeline Architecture v.4.} \label{iot}
\end{figure}
  In raspberry pi, we use local lambda functions to deploy functions for the sensor and ML models. We use Greengrass to manage lambda function deployment. In AWS cloud, we pair IoT analytic with Quicksight so we can check on the new data samples. We found that it is more efficient to preprocess data in the Sagemaker pipeline. So in the coming data sample is put directly into the DynamoDB table and backed up to a S3 bucket. When the Sagemaker pipeline produces an approved anomaly detection model, we can use Greengrass to deploy a local lambda function in raspberry pi. If the anomaly detection model in raspberry pi detects an anomaly, it will send a message to the IoT Core in AWS. The IoTCore will then process the message, invoke the lambda function, and send a notification email with the anomaly data samples. 
  
  For the modeling part, it uses real-world data generated from the sensor modules, which include a temperature sensor, pressure sensor, lux sensor, gyroscope, RGB sensor, acceleration sensor, and Amazon DynamoDB as the database. Data from various sensor modules are uploaded to the DynamoDB database. In the Python script, all the data stored are retrieved, preprocessed, the anomaly detection model and the clustering model are trained, and the results are analyzed. The whole pipeline is handled by Amazon Sagemaker. 

\section{Implementation}
\subsection{Data Generation}
  In this project, we used the WaveShare Sense HAT(B)’s onboard sensors to collect environmental data. After the data is collected, the software will first store these samples in a local SQL database as cache. For every ten seconds, the raspberry pi will connect to the AWS cloud and send the data to the AWS cloud using MQTT messages. 
\subsubsection{WaveShare Sense HAT(B)}
  WaveShare Sense HAT(B) \cite{r24} is a prototyping board designed for raspberry pi. It integrated several sensors we used to collect environmental data for the project. Raspberry pi can communicate with the HAT using the Inter-Integrated Circuit (I2C) interface, the HAT also allows add-on analog and I2C sensors if needed. Table 1 shows the available sensors and their specifications. 
  
\begin{table}[]
\caption{The specifications of the available sensors of WaveShare Sense HAT(B)}
\scalebox{0.80}{
\begin{tabular}{|l|l|l|}
\hline
\textbf{Sensor Type}        & \textbf{Model Name} & \textbf{Specification}                                                                                                                                                              \\ \hline
Gyroscope                   & ICM-20948           & \begin{tabular}[c]{@{}l@{}}Ranging: ±250/500/1000/2000 dps\\    \\ Resolution: 16-bit\end{tabular}                                                                                  \\ \hline
Accelerometer               & ICM-20948           & \begin{tabular}[c]{@{}l@{}}Ranging: ±2/4/8/16 g\\    \\ Resolution: 16-bit\end{tabular}                                                                                             \\ \hline
Magnetometer                & ICM-20948           & \begin{tabular}[c]{@{}l@{}}Ranging: ±49 gauss\\    \\ Resolution: 16-bit\end{tabular}                                                                                               \\ \hline
Barometer                   & LPS22HB             & \begin{tabular}[c]{@{}l@{}}Ranging: 260 - 1260 hPa\\    \\ Accuracy (ordinary temperature): ±0.025hPa\\    \\ Speed: 1 Hz - 75 Hz\end{tabular}                                      \\ \hline
Temperature \& Humidity      & SHTC3               & \begin{tabular}[c]{@{}l@{}}Accuracy (Humidity): ±2\% rH\\    \\ Ranging (Humidity): 0\% ~ 100\% rH\\    \\ Accuracy (Temp): ±0.2°C\\    \\ Ranging (Temp): -30 ~ 100°C\end{tabular} \\ \hline
Color Sensor                & TCS34725            & High precision 12-bit ADC                                                                                                                                                           \\ \hline
Analog to Digital Convertor & LSF0204PWR          & 12 bit ADC                                                                                                                                                                          \\ \hline
\end{tabular}
}
\end{table}

\subsubsection{Sensor Read/Cache/Send Software}
  The software that reads, stores, and sends to AWS Cloud is containerized in a lambda function and deployed to each raspberry pi using Greengrass. The software was originally provided by the author  “janrutger” in his GitHub repository \cite{r25}. The software is modified to better integrate it into our project. 
  
  When the local lambda function is triggered by Greengrass, it will first initialize the connection to the AWS IoT Core.  Then, it will configure the Waveshare Sense HAT via the I2C interface and create an empty SQLite database as a cache. When the configuration is finished, it enters a loop and starts reading sensor data. Every second, the software will read sensors specified in the configuration and store them in the cache database as one sample entry. Every ten seconds, the software will connect to AWS IoT Core and send those new samples as MQTT messages.
\subsection{Data Transmission}
  To successfully send a MQTT message to the IoT Core, Greengrass in the raspberry pi will need a root-CA certificate, a X.509 client certificate and an asymmetric private key. When Greengrass sends a connection request to the AWS IoT Core server, the server will reply with a X.509 signed certificate. The software will compare the X.509 provided certificate and the one returned from the server to verify that it is connecting to the correct server, not a server impersonating AWS IoT Core. An additional client certificate is needed to verify Greengrass in raspberry pi is connected to the correct “things” registered in AWS IoT Core. When Greengrass sends a MQTT message to AWS IoT Core, it will use the private key to encrypt the MQTT message and send it as a HTTP packet. 
\subsection{Data retrieval}
\subsubsection{PostgreSQL}
  For the PostgreSQL database, the connection is established using the psycopg2 library. A cursor is created to run a SQL query that returns all the rows in the table. They are fetched and transformed into a NumPy array. After saving all the column names in a list, it is combined with the NumPy array and changed into a pandas dataframe. Then the connection and cursor are closed to save resources.
\subsubsection{DynamoDB}
  The package boto3 is used to connect to DynamoDB. After connecting to the database, it will try to increase the provisioned read throughput to 500 units in order to speed up reading data from the database. Then, the table is scanned to send back data. The scan function will return a batch of data of size 1MB with a ‘LastEvaluatedKey’. In the loop, it will continuously extend the results list with new batches of data until there is no more data (‘LastEvaluatedKey’ is null), and exits the loop.
\subsection{Data Preprocessing}
\subsubsection{Sample dataset}
  The sample dataset downloaded from DynamoDB includes data gathered from multiple sensors with anomaly labels that can be used for supervised learning. The goal is to test different models on the sample data with a focus on their performances and speeds as well as the effects on performance after applying the Synthetic Minority Over-sampling Technique (SMOTE) on the anomaly data. 
  
  SMOTE \cite{r26} is one of the methods to overcome the issue of class-imbalanced data. It can create more instances for the minority classes to balance the dataset based on the idea of K-nearest Neighbors \cite{r27}. First, a random sample from the minority class is selected. Then k nearest neighbors are found and one of them is selected as the second basis. A new instance is created randomly between the two points.Thus,  SMOTE is applied to test if it can improve the performance of models \cite{r28}.
  
  As Figure 5 indicates, the sample dataset has a total of 357,952 entries and 13 columns. Only column ‘timestamp’ is of type int64, while all other columns are strings.
 
  \begin{figure}
     \centering
     \includegraphics[width=8cm]{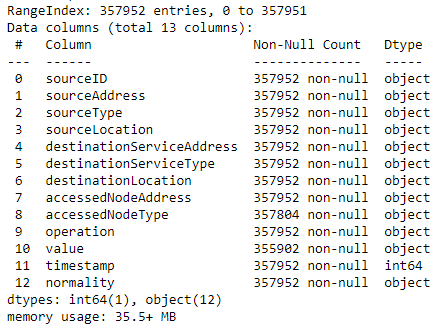}
     \caption{Data information of the sample dataset.} \label{iot}
\end{figure}

  There are a small number of null values in the ‘accessedNodeType’ and  ‘value’ columns. The first step of preprocessing is to fill all the null values and then transform string features into numeric values through label encoding \cite{r29}. Based on findings and patterns of the data, the null values of ‘accessedNodeType’ column are filled with ‘/batteryService’ and the null values of ‘value’ column are filled with 0. The rest of the columns are encoded into integers that start from 0. For normality values, normal values are encoded to 0 and abnormal values are encoded to 1. All the encoded columns will be the features and normality will be the label used for prediction.
  
\subsubsection{Real sensor data}
  The sensor data uploaded to DynamoDB has 21 columns and approximately 2.8 million entries (as of time of writing) in total, as shown in Figure 6. Each column represents a value gathered from a sensor. Sensors such as gyroscope, RGB sensor, acceleration sensor, and magnetic sensor generate three different values instead of one. In this case, each value takes a column. 
  
  \begin{figure}
     \centering
     \includegraphics[width=8cm]{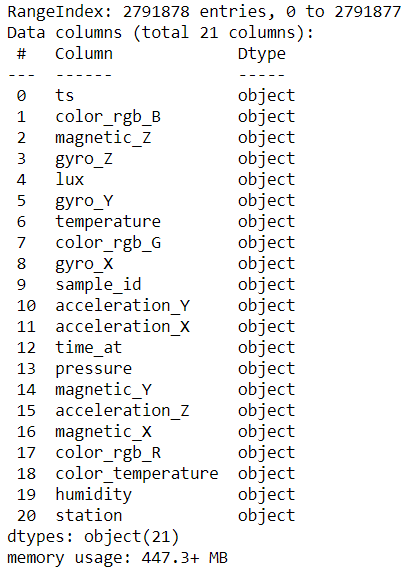}
     \caption{Data information of the real sensor dataset.} \label{iot}
\end{figure}

  The preprocessing stage includes the following steps:
\begin{enumerate}
\item Fill all the null values with 0.
\item Search for duplicate entries with the same ‘sample\_id’ and remove them from the DataFrame as well as DynamoDB.
\item Remove samples where ‘magnetic\_X’,  ‘magnetic\_Y’ and  ‘magnetic\_Z’ are all 0.
\item Encode ‘station’ column: 'daniel/house/garage/pi3' to 0,\\ 'daniel/house/bedroom/pi4' to 1. Drop all other stations.
\item Convert ‘time\_at’ column from string to Unix timestamp/Epoch. 
\item Drop the ‘ts’ column.
\item Reset index of the dataframe.
\end{enumerate}

  The preprocessing stage cleans and prepares the data for modeling and analysis in the next step. The analysis was done separately for two different environments, which are garage and bedroom.
  
\subsection{Data processing and modeling}
\subsubsection{Sample dataset}
  Three models and SMOTE oversampling are used for the purpose of result comparison. Models include Extreme Gradient Boost (XGBoost) \cite{r30} and Random Forest (RF) \cite{r31} algorithm. 
  
  XGBoost is a gradient boosting decision tree algorithm designed for computational speed and model performance \cite{r30}. It is optimized through parallel processing, tree-pruning, handling missing values, and regularization to prevent overfitting \cite{r32}. Its hyperparameters include the number of estimators, maximum depth, learning rate, and type of objective.  RF is a bagging ensemble learning algorithm that uses only a subset of features selected randomly to build the forest or decision trees \cite{r31}. Besides the number of estimators, maximum depth, its hyperparameters also include maximum features, minimum samples split, minimum samples leaf, and criterion \cite{r28}. 
  
  All models are tuned by Bayesian Optimization with Tree-structured Parzen Estimator (BO-TPE). BO-TPE creates two density functions, l(x) and g(x) that indicate the probability of finding the next hyperparameter in well-performing and poor-performing regions. It finds the optimal hyperparameter combination by maximizing the value of $l(x)/g(x)$ \cite{r33}.
  
  The training and testing set ratio is set to 7:3. The evaluation metrics use accuracy, precision, recall, and f1 score to show the performance of the model. The metrics are based on True Positive (TP), True Negative (TN), False Positive (FP), and False Negative (FN) values as below \cite{r28}. 
  
\begin{equation}
Acc= \frac{T P+T N}{T P+T N+F P+F N} 
\end{equation}
\begin{equation}
Precision=\frac{T P}{T P+F P} 
\end{equation}
\begin{equation}
Recall=\frac{T P}{T P+F N} 
\end{equation}
\begin{equation}
F 1=\frac{2 \times T P}{2 \times T P+F P+F N} 
\end{equation}

\subsubsection{Real sensor data}
  The model used for anomaly detection is Isolation Forest \cite{r12}, a tree-based algorithm that recursively selects a random feature and splits it by a random value. It uses the idea that anomalies are easier to be isolated from the entire set. The algorithm will return either 1 or -1, which represents normal value and outlier respectively. 
  
  For clustering the anomalies, K-means is used as the model. It is one of the most popular clustering algorithms and unsupervised learning algorithms. It starts by randomly selecting centroids as the initial center for each cluster. Then it repetitively calculates the distances of the data points, assigns them to different clusters, and optimizes the location of centroids until all the centroids have stabilized or the defined number of iterations has been achieved \cite{km}. 
  
  The training and testing set ratio is set to 7:3, and the anomaly rate is set to an estimate of 5\%. The results are plotted into several graphs to visually evaluate the performance of the model.

\section{Results and Discussion}
The data analysis and modeling results can be divided into two stages in correspondence to two different versions of the data processing architecture. The first stage intends to analyze a sample dataset stored in PostgreSQL database by applying two different supervised learning algorithms (XGBoost and Random Forest) and SMOTE oversampling for result comparison. Confusion matrix and four scores (accuracy, precision, recall, f1 score) are used to evaluate the performance of both models. The second stage uses data collected from two sets of sensors located in two different environments, a bedroom and a garage. It aims to identify special cases such as an individual working in the bedroom and opening the garage door through anomaly detection and K-means clustering. 
\subsection{Sample dataset}
  This part displays and compares the results of applying XGBoost and Random Forest before and after using SMOTE to oversample the minority cases or anomalies. 
  
  The following four figures are the confusion matrix and the four scores of applying XGBoost and Random Forest models without using SMOTE algorithm to increase the minority cases, which are the anomalies in this case. 
  
  The confusion matrix visualizes the results of classification by displaying two correct classification cases, True Positive and True Negative, as well as two incorrect classification class cases, False Positive (Type I error) and False Negative (Type II error). As the result shows, Random Forest performed slightly better than XGBoost and is higher in all four scores by less than 0.01\%. Both two models produce more Type I errors than Type II errors.

  \begin{figure}
     \centering
     \includegraphics[width=8cm]{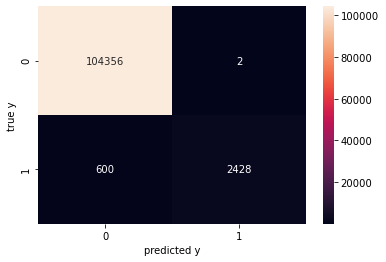}
     \caption{Confusion matrix of XGBoost (without SMOTE).} \label{iot}
\end{figure}

\begin{table}[]
\centering%
\setlength\extrarowheight{1pt}
\caption{Four scores of XGBoost (without SMOTE)}
\begin{tabular}{|l|l|}
\hline
Accuracy   & 0.994394 \\ \hline
Precision  & 0.994421 \\ \hline
Recall     & 0.994394 \\ \hline
F1   score & 0.994094 \\ \hline
\end{tabular}
\end{table}

  \begin{figure}
     \centering
     \includegraphics[width=8cm]{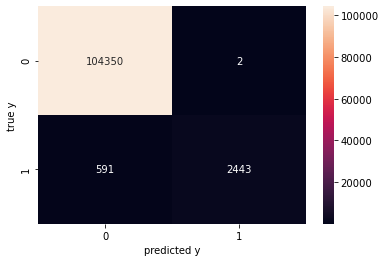}
     \caption{Confusion matrix of RF (without SMOTE).} \label{iot}
\end{figure}

\begin{table}[]
\centering%
\setlength\extrarowheight{1pt}
\caption{Four scores of RF (without SMOTE)}
\begin{tabular}{|l|l|}
\hline
Accuracy   & 0.994478 \\ \hline
Precision  & 0.994504 \\ \hline
Recall     & 0.994478 \\ \hline
F1   score & 0.994189 \\ \hline
\end{tabular}
\end{table}

  The following figures show the confusion matrix and the four scores of the two models after applying SMOTE, which oversamples the number of anomalies from approximately 7,000 to 15,000. 15,000 was selected for result comparison as it produces the best results after manually testing with other numbers.
  
  Both models' accuracy, recall, and f1 score decrease after oversampling and the error type switches from Type I to Type II. Random Forest still performs slightly better than XGBoost. It is higher by around 0.03\% in all scores and has 41 fewer error cases.

  \begin{figure}
     \centering
     \includegraphics[width=8cm]{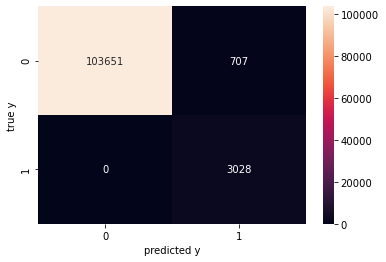}
     \caption{Confusion matrix of XGBoost (after SMOTE).} \label{iot}
\end{figure}

\begin{table}[]
\centering%
\setlength\extrarowheight{1pt}
\caption{Four scores of XGBoost (after SMOTE)}
\begin{tabular}{|l|l|}
\hline
Accuracy   & 0.993416 \\ \hline
Precision  & 0.994663 \\ \hline
Recall     & 0.993416 \\ \hline
F1   score & 0.993749 \\ \hline
\end{tabular}
\end{table}

  \begin{figure}
     \centering
     \includegraphics[width=8cm]{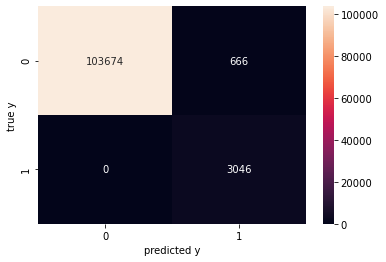}
     \caption{Confusion matrix of RF (after SMOTE).} \label{iot}
\end{figure}

\begin{table}[]
\centering%
\setlength\extrarowheight{1pt}
\caption{Four scores of RF (after SMOTE)}
\begin{tabular}{|l|l|}
\hline
Accuracy   & 0.993798 \\ \hline
Precision  & 0.994911 \\ \hline
Recall     & 0.993798 \\ \hline
F1   score & 0.994094 \\ \hline
\end{tabular}
\end{table}

  In general, the Random Forest algorithm produces better results in both cases. However, XGBoost is more computationally efficient and is much faster than RF, especially when running the tuning algorithm. After applying SMOTE oversampling, the errors switch from Type I to Type II errors. In this case, having Type II errors is more preferable to having Type I errors as being unable to detect anomalies may result in serious problems.

\subsection{Real sensor data}
\subsubsection{Bedroom environment}
  The bedroom subset has a total of 626,865 entries which were gathered over a period of days with one individual living and working inside. The sensors record the changes in values that were possibly caused by turning on/off light, temperature/pressure/humidity changes, touching or moving Raspberry Pi, sensor anomalies, and etc. Figure 11 shows the result of applying the anomaly detection model, Isolation Forest. Red represents anomalies, and blue represents normal data points. There are 595,521 normal values and 31,344 abnormal values. The abnormal and normal data points are not clearly separated into different clusters.

  \begin{figure}
     \centering
     \includegraphics[width=8cm]{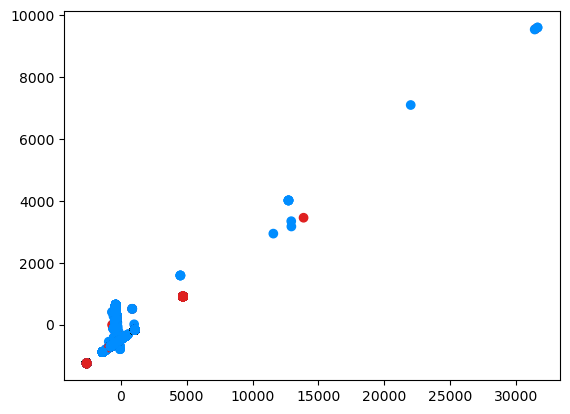}
     \caption{Result of Isolation Forest on bedroom data.} \label{iot}
\end{figure}

  \begin{figure}
     \centering
     \includegraphics[width=13cm]{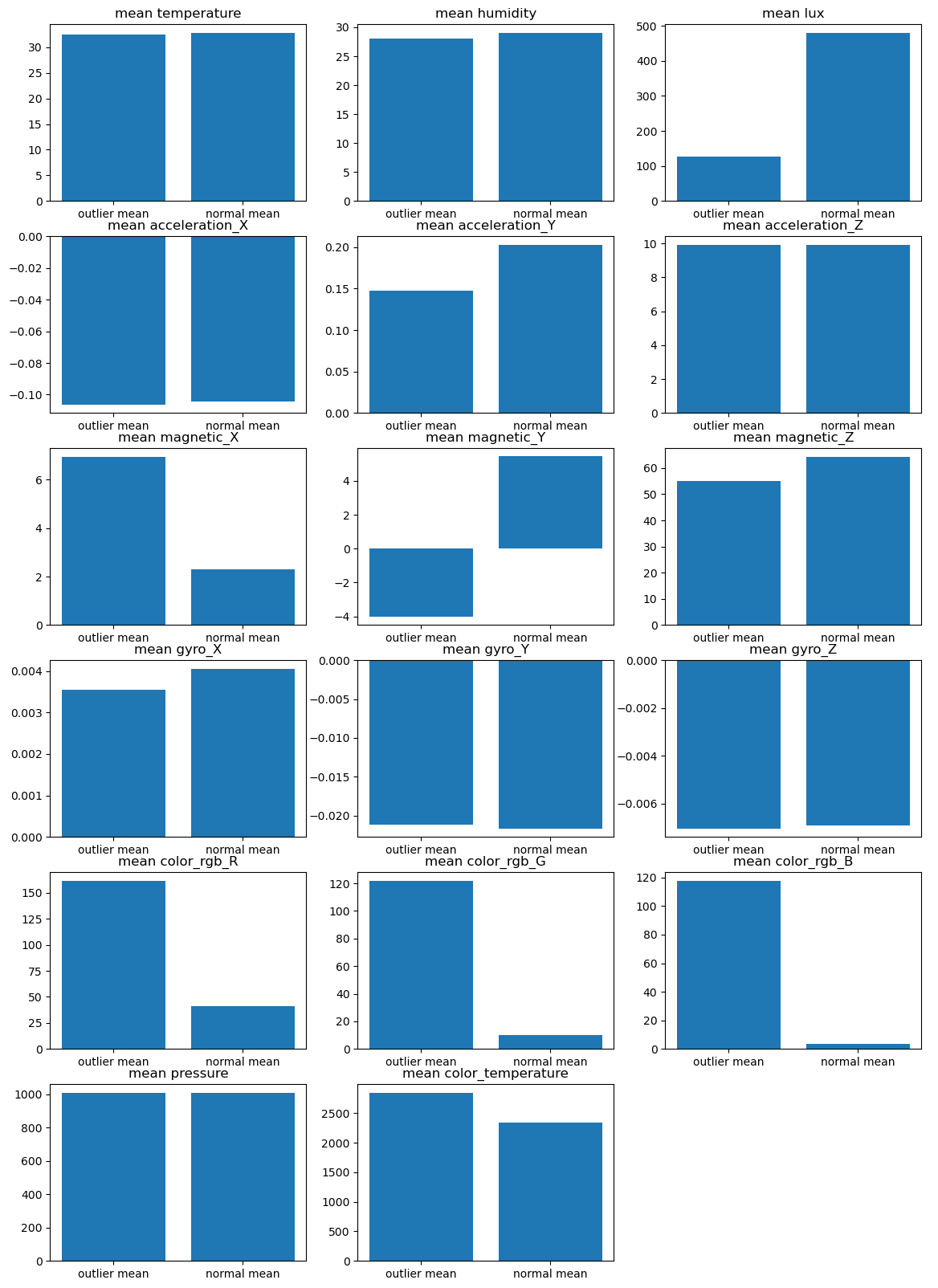}
     \caption{Abnormal vs. Normal mean feature values (bedroom).} \label{iot}
\end{figure}

  According to Figure 12, some features show an apparent difference between abnormal and normal values including lux, acceleration Y, magnetic X, magnetic Y, all RGB values, and color temperature. The normal values may represent the bedroom in a normal condition with a medium level of lighting, temperature, and humidity.
  
  To further analyze the anomaly data, they are extracted and clustered into 4 groups using the K-means algorithm. Each cluster is represented by a unique color, as Figure 13 shows.

  \begin{figure}
     \centering
     \includegraphics[width=8cm]{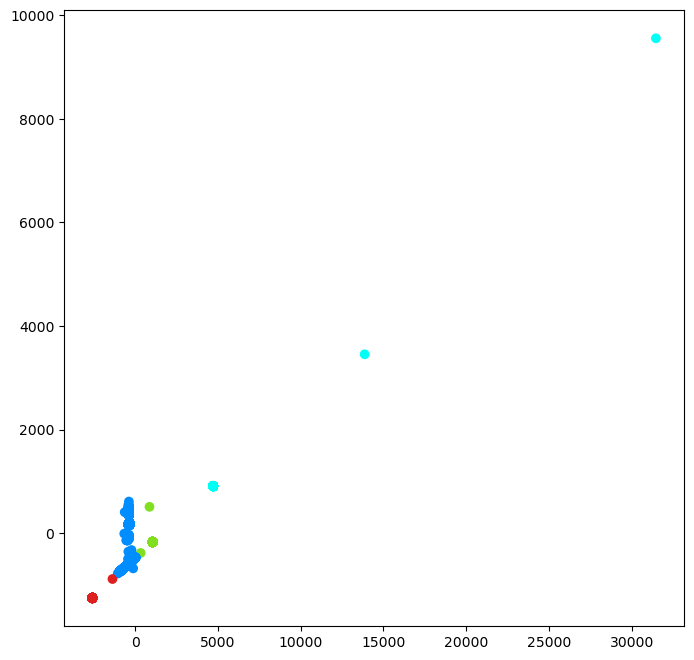}
     \caption{Result of K-means clustering on anomalies (bedroom).} \label{iot}
\end{figure}

  Figure 14 displays the mean feature values of each cluster of anomalies as well as the normal mean value. It is obvious that cluster 2 is different in many features, besides acceleration, gyroscope, and magnetic anomalies that could be caused by touching or moving the device. There are large differences in the mean temperature and lux values. Cluster 2 has a high temperature and lux level compared to normal, which can be caused by the individual working in the bedroom with lights and computer screen on. A running computer could release a large amount of heat, and closing the bedroom door could retain the heat causing the mean temperature to increase. This case of an individual working in the bedroom can be a confirmation that there are people at home and use for the anti-theft security system.

  \begin{figure}
     \centering
     \includegraphics[width=12cm]{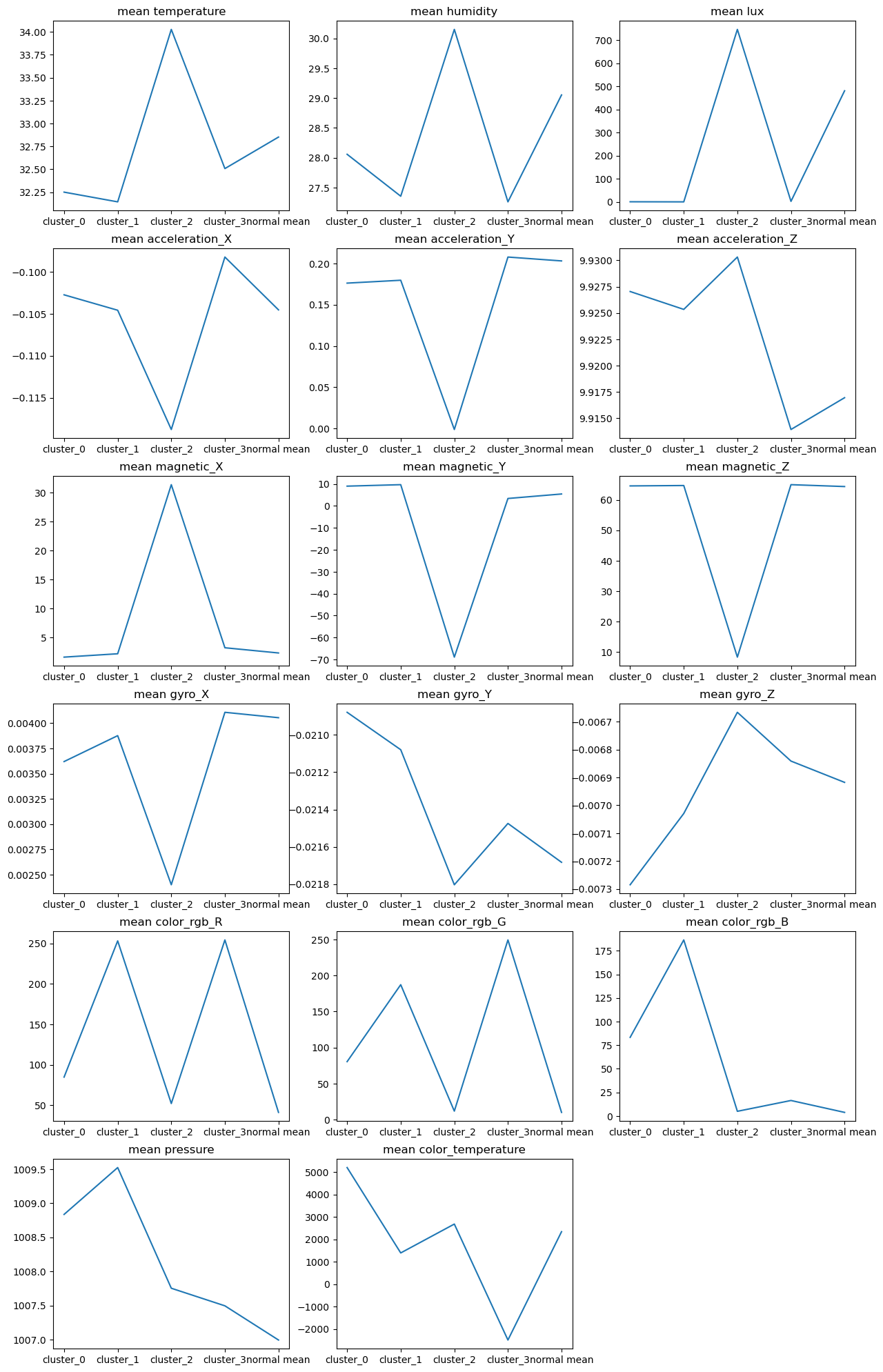}
     \caption{Mean feature values of each cluster and normal mean value (bedroom).} \label{iot}
\end{figure}
 
   \begin{figure}
     \centering
     \includegraphics[width=10cm]{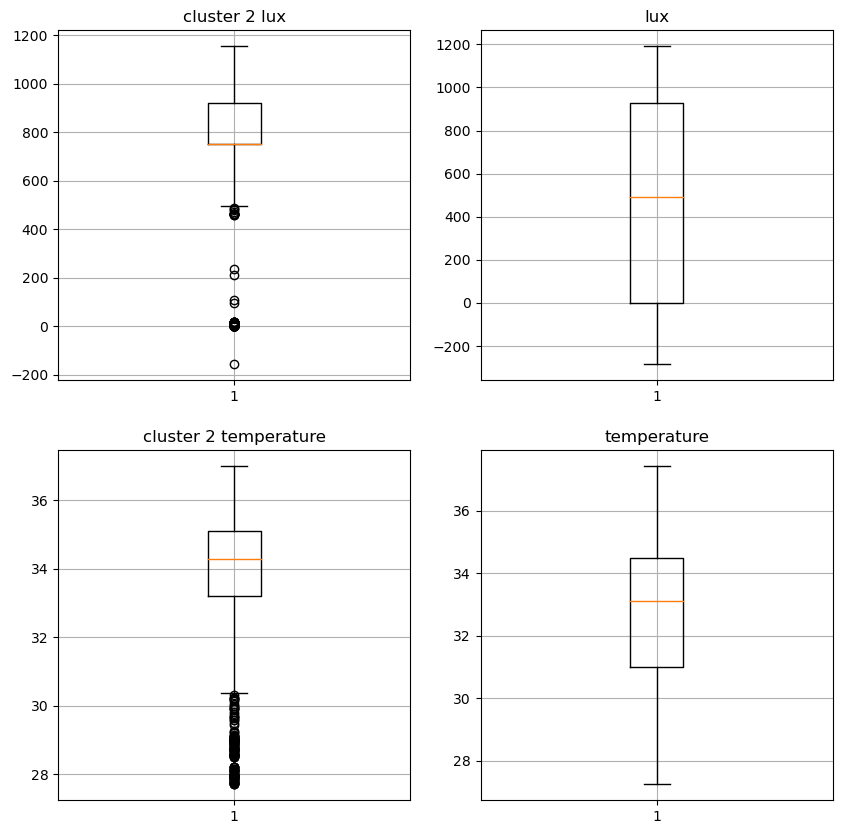}
     \caption{Boxplots of cluster 2 lux and temperature values vs. normal values (bedroom).} \label{iot}
\end{figure}

\subsubsection{Garage environment}
  For the garage subset of data, there are a total of 1,119,942 rows. Different from the bedroom environment, for the majority of the time, the garage remains dark. However, there are cases where people enter the garage, turn on lights, open the garage door, drive the vehicle out, and people practicing guitar in the garage with lights on. These cases could all cause fluctuations and may be the cause of anomalies. 
  
  Figure 16 shows the result of the Isolation Forest on garage data. There are a total of 1,063,944 normal values and 55,998 anomalies. Normal values gather at the bottom part of the graph while anomalies take the top section.
 
    \begin{figure}
     \centering
     \includegraphics[width=8cm]{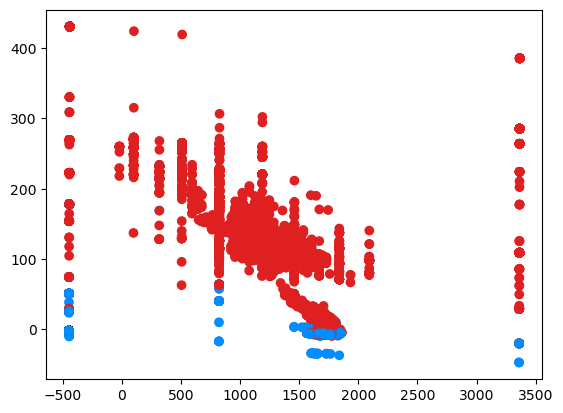}
     \caption{Result of Isolation Forest on garage data.} \label{iot}
\end{figure}

  As shown in Figure 17, the abnormal mean values of lux and RGB are very different from normal means values. It may simply mean that the garage is of low lux level for the majority of the time. Any activities in the garage may cause the lux level to increase, for example, opening the garage door and people turning on the lights in the garage. RGB values and color temperature also vary in accordance with the lux value.
 
     \begin{figure}
     \centering
     \includegraphics[width=13cm]{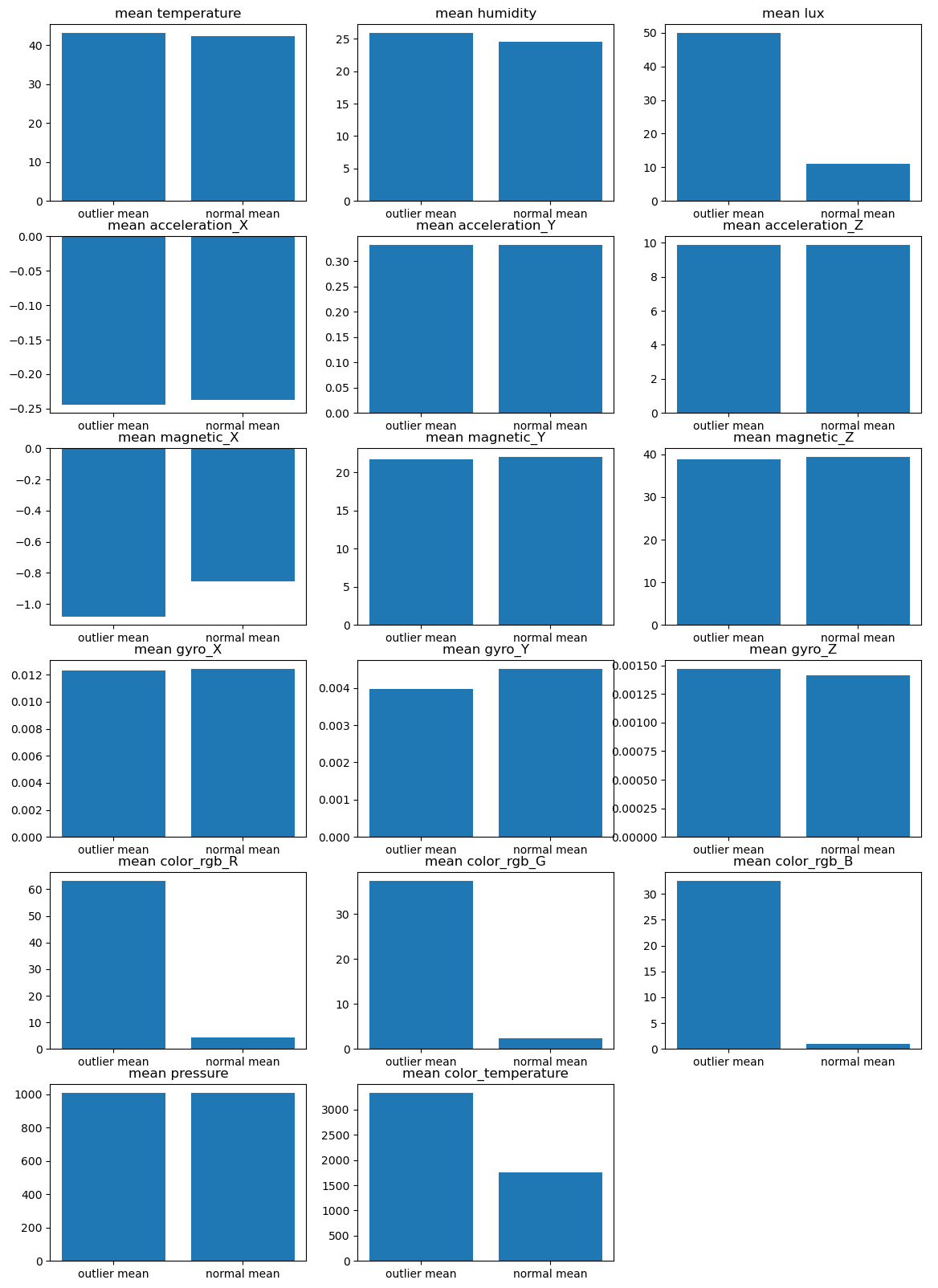}
     \caption{ Abnormal vs. Normal mean feature values (garage).} \label{iot}
\end{figure}

    \begin{figure}
     \centering
     \includegraphics[width=8cm]{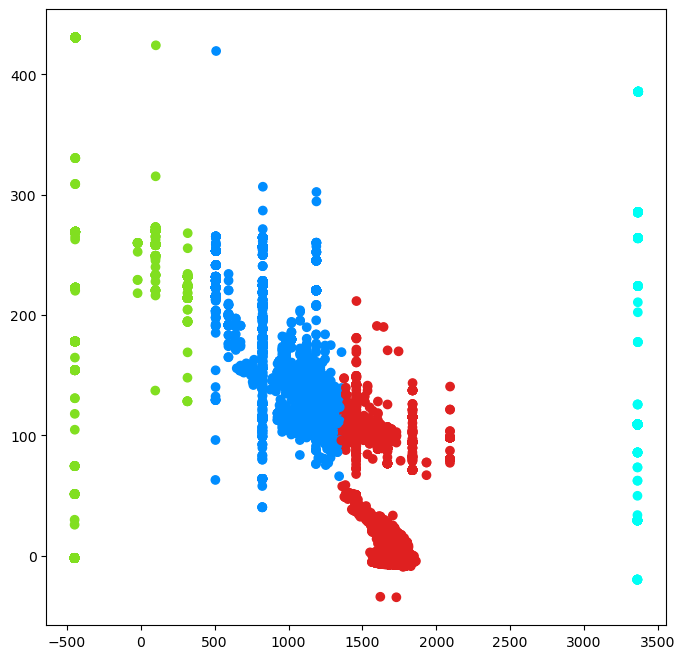}
     \caption{Result of K-means clustering on anomalies (garage).} \label{iot}
\end{figure}

     \begin{figure}
     \centering
     \includegraphics[width=11.5cm]{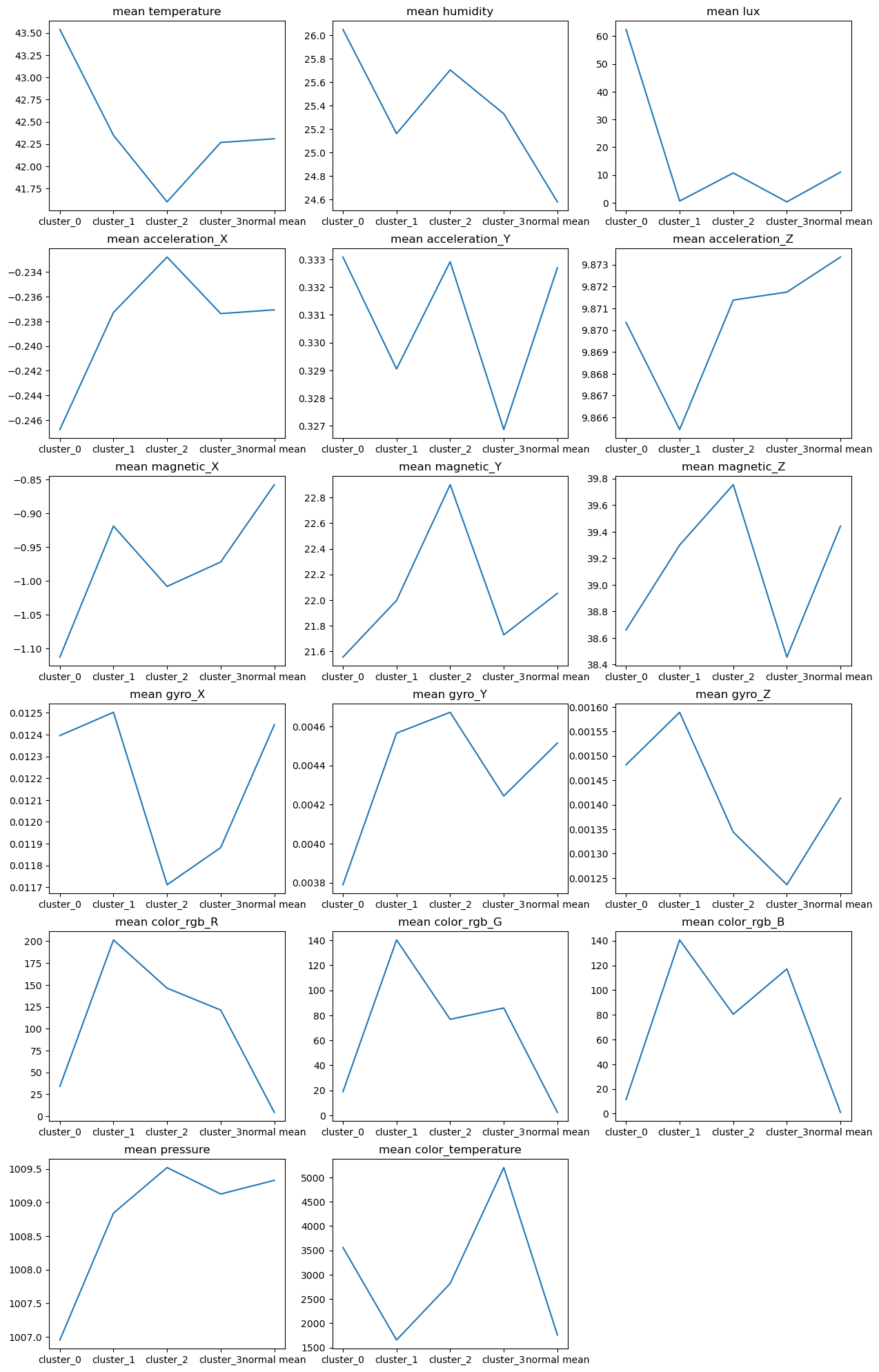}
     \caption{Mean feature values of each cluster and normal mean value (garage).} \label{iot}
\end{figure}
 
     \begin{figure}
     \centering
     \includegraphics[width=8cm]{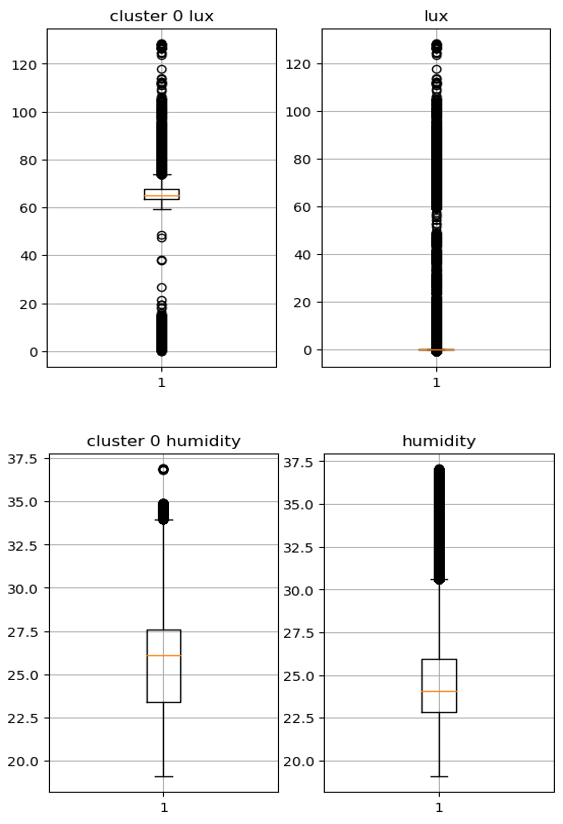}
     \caption{Boxplots of cluster 0 lux and humidity values vs. normal values (garage).} \label{iot}
\end{figure}

  In order to look into the details of anomalies, the K-means model is applied to differentiate them into 4 clusters after leaving out the normal values, as shown in Figure 18.

  Figure 19 contains the plots of all mean feature values of each cluster compared to the normal mean values. The changes in acceleration, magnetic, and gyroscope are not large compared to the scales of other features. It is worth noting that the temperature, humidity, and lux of cluster 0 are different from normal values and other clusters. They all have higher values than normal. Due to the fact that only turning on lights in the garage would not change the temperature and humidity, the combination of higher levels in all three features may be caused by the opening of the garage door. The temperature and humidity inside the garage may change slightly when the air outside enters the garage.

  When combining the cases of both the bedroom and the garage, it becomes the situation where the individual is working in the bedroom but the garage door opens. This can be considered as an anomaly different from the normal pattern, and also, a case for the anti-theft security system that protects the home from outside intrusion.

\section{Conclusion}
To ease the process of IoT anomaly detection, this work proposed an automated data pipeline for IoT sensor anomaly detection that is used for the anti-theft home security system. The project first uses a sample dataset for model testing. After preprocessing, results from XGBoost and Random Forest, before and after SMOTE oversampling, are compared based on the confusion matrix and four evaluation scores (accuracy, precision, recall, f1 score). The second stage of the project focuses on building the data pipeline, which starts by uploading real-time sensor data from two sets of sensor modules connected to two Raspberry Pi machines. It utilizes multiple Amazon Web Services to automate the tasks containing different models, which detects anomalies with Isolation Forest, analyzes different anomalous clusters, and finally identifies a cluster representing a special case of home intrusion where the garage door opens while having individuals working in the bedroom.

\section{Acknowledgement}
We would like to acknowledge and thank Prof. Abdallah Shami and Mr. Li Yang from OC2 Lab, Department of Electrical and Computer Engineering, Western University, for their support and help throughout this work.

\end{document}